\documentclass[letterpaper, 10 pt, conference]{ieeeconf}  %

\IEEEoverridecommandlockouts                              %

\usepackage{amsmath} %
\usepackage{amssymb}  %
\usepackage{multirow}  %
\usepackage{xcolor}    %
\usepackage{ragged2e}  
\usepackage{graphicx}
\usepackage{svg}
\usepackage{array}
\usepackage{booktabs}       %
\usepackage{threeparttable} %
\usepackage{bbm}
\newcolumntype{C}[1]{>{\centering\arraybackslash}p{#1}}
\newcolumntype{R}[1]{>{\raggedleft\arraybackslash}p{#1}}
\usepackage{cite}

\title{\LARGE \bf
CoGaussian: Open-Vocabulary Gaussian-based Collaborative \\ 3D Scene Understanding
}

\author{Om Shivam Verma$^{1*}$, Shaya Farahmand$^{1*}$, Lingjun Zhao$^{1}$, Lu Gan$^{1}$%
\thanks{*These authors contributed equally to this work.}%
\thanks{$^{1}$Lunar Lab, Georgia Institute of Technology,
        Atlanta.
        {\tt\small \{overma3,sfarahmand3,lzhao360,lgan\}@gatech.edu}}%
}

\begin{document}

\maketitle
\thispagestyle{empty}
\pagestyle{empty}

\begin{abstract}

Collaborative perception enables agents to retrieve the scene semantics and geometry beyond the field of view of any individual agent. This broader scene understanding enhances transportation safety and improves efficiency in multi-robot systems. However, existing collaborative semantic occupancy methods rely on dense 3D voxel annotations and are limited to recognizing objects within a predefined set of semantic classes. To overcome these limitations, we present CoGaussian, an open-vocabulary Gaussian-based collaborative semantic occupancy prediction framework. CoGaussian represents a scene as a set of foundation model-aligned sparse 3D Gaussians, which serve as a compact representation for communicating and fusing geometric and semantic information across multiple agents. On the Semantic-OPV2V benchmark, our method outperforms existing open-vocabulary single-agent approaches, highlighting the effectiveness of open-vocabulary collaborative scene understanding.
\end{abstract}

\section{INTRODUCTION}

Collaborative perception enables local agents to exchange information that may be occluded or beyond their field-of-view, improving geometric and semantic understanding of their environment, and enabling safer and more reliable decision making in autonomous driving and robot navigation. Semantic occupancy prediction (SOP) is a representative and challenging task in 3D spatial understanding, which aims to predict voxel-wise occupancy status and semantic class. Recently, 3D Gaussian-based methods~\cite{3dgs} have been shown to be effective for SOP. The GaussianFormer family~\cite{gaussianformer, gaussianformer2, zhao2025gaussianformer3d, gaussianworld} represents scenes as 3D Gaussians, supervised by dense occupancy labels. VOGS-CP~\cite{vogs-cp} extends GaussianFormer to collaborative SOP which communicates Gaussian primitives between neighboring agents. However, these methods are limited to predicting semantics within a closed vocabulary and heavily rely on densely annotated 3D occupancy labels that are expensive to obtain.

Vision foundation models (VFMs), such as DINO~\cite{dinov2} and CLIP~\cite{clip} have demonstrated strong open-vocabulary capabilities in 2D perception. In addition, Talk2DINO~\cite{barsellotti2025talkingdinobridgingselfsupervised} and DINO.txt~\cite{dinotxt} further align 2D VFM features with natural language. 
Building on the above VFMs, GaussTR~\cite{gausstr} represents scenes as a set of sparse open-vocabulary 3D Gaussians that are predicted in a feed-forward manner and can be queried via natural language. The Gaussians are encoded with geometry and semantic information by splatting into 2D views and aligning rendered features with foundation model-derived depth~\cite{metric3dv2} and semantics~\cite{clip, dinov2}. However, GaussTR is designed for single-agent perception and cannot leverage information that is visible only to neighboring agents.

Based on these observations, we propose \textbf{CoGaussian}, the first open-vocabulary multi-robot collaborative perception framework. We represent scenes as sets of sparse VFM-aligned 3D Gaussians, which also serve as the communication primitives among agents. We also explore and develop three different algorithms to aggregate and fuse information from neighboring agents. Our contributions can be summarized as follows:

\begin{itemize}
    \item \textbf{Open-Vocabulary Collaborative Perception:} We propose CoGaussian, the first framework for open-vocabulary collaborative perception (e.g. 3D semantic occupancy prediction) in outdoor environments.
    
    \item \textbf{Multi-Robot 3D Gaussian Representation:} We develop a foundation model-aligned 3D Gaussian representation for collaborative perception, along with fusion strategies to enhance information transfer among agents.
    
    \item \textbf{Zero-Shot Benchmarking:} We evaluate CoGaussian on the Semantic-OPV2V~\cite{opv2v,cohff} dataset in a \textit{zero-shot} manner. Without any task-specific supervision, our approach achieves impressive performance for semantic occupancy prediction, highlighting the potential for open-vocabulary collaborative perception.
\end{itemize}

\begin{figure}
    \centering
    \includegraphics[width=0.5\textwidth]{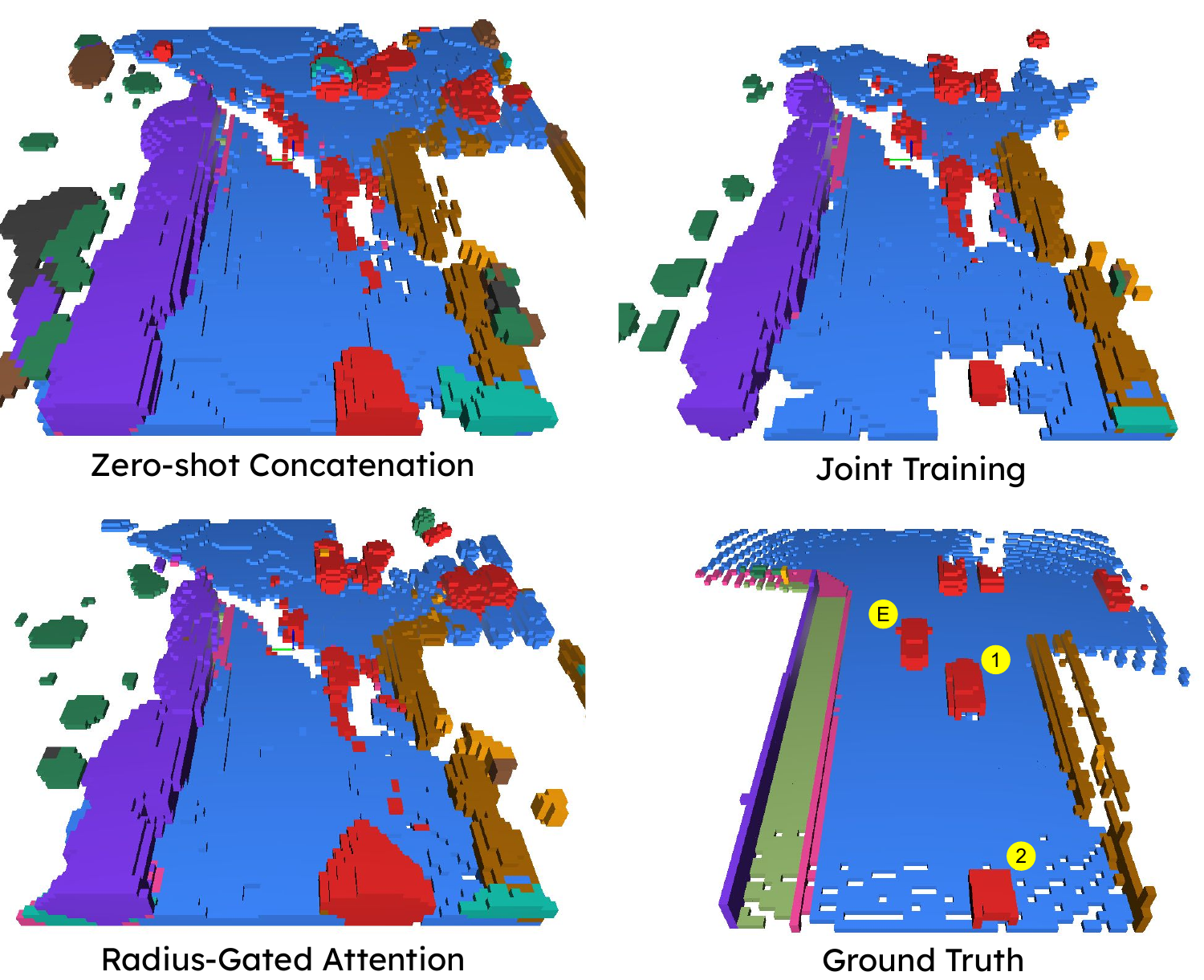}
    \caption{\textbf{Qualitative comparison of three variants of CoGaussian with ground truth.} The ego vehicle is marked as E. Another agent is in the front-right (not visible here), partially completing the front-right of the ego vehicle in collaborative predictions. Class legends are shown in Table~\ref{tab:main}.}
    \label{fig:visualisation}
\end{figure}

\begin{figure*}
    \centering
    \includegraphics[width=\textwidth]{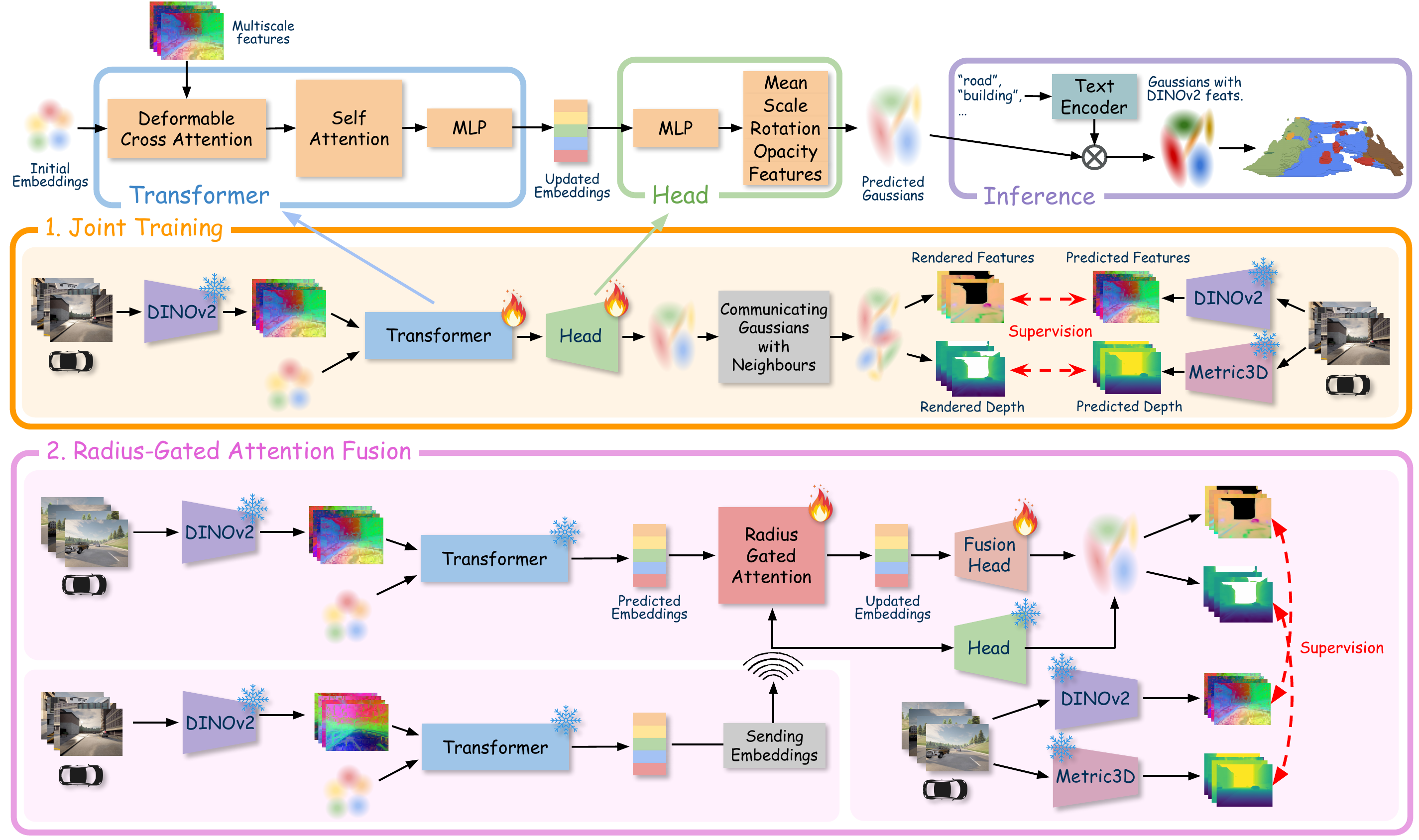}
    \caption{\textbf{Overview.} CoGaussian starts with randomly initialized embeddings. A transformer layer uses multiscale features of images to update the embeddings, followed by a Head Layer that predicts Gaussians. During Joint Training, all agents predict \textit{ego Gaussians} which are communicated with each other and aggregated. This set of Gaussians is used for training. In the Fusion pipeline, the neighbors predict and send Gaussian embeddings to the ego agent. Radius-gated Attention uses these \textit{neighbor Gaussian} embeddings to update ego embeddings. The Fusion Head predicts ego Gaussians, and pre-trained Head from Joint Training predicts neighbor Gaussians. These are aggregated to form the final prediction used for both training and inference.}
    \label{fig:method-arch}
\end{figure*}

\section{Method}
\label{sec:method}

We represent a scene as a sparse set of 3D Gaussians with associated DINOv2~\cite{dinov2} features. Predicted Gaussians are splatted back into the source camera view and supervised by DINOv2~\cite{dinov2} features and Metric3Dv2~\cite{metric3dv2} monocular depth. During inference, semantic class logits are computed as similarity between the predicted Gaussian features and Talk2DINO~\cite{barsellotti2025talkingdinobridgingselfsupervised} text embeddings for each class.

\subsection{Preliminaries}
\label{sec:preliminaries}
Each 3D Gaussian $G$ is parametrized as:
$$
G = \{\boldsymbol{\mu},\,
        \mathbf{s},\,
        \mathbf{r},\,
        \alpha,\,
        \mathbf{f} \}, \eqno{(1)}
$$
where $\boldsymbol{\mu} \in \mathbb{R}^{3}$ is the mean (center), $\mathbf{s} \in \mathbb{R}^{3}$ is the per-axis scale, $\mathbf{r} \in \mathbb{R}^{4}$ is the rotation quaternion of the axes, $\alpha \in [0,1]$ is the opacity, and $\mathbf{f} \in \mathbb{R}^{D}$ is the high-dimensional DINOv2 feature vector. 
The model inputs an image and initializes a set of Gaussian embeddings $\{\mathbf{q_i} \in \mathbb{R}^E\}_{i=1}^{N}$, where $E$ denotes the dimension of the embedding. Each embedding vector is associated with a randomly sampled location $(u_i, v_i) \in [0,1]^2$ on the image and its corresponding interpolated Metric3D depth value. The coordinates are concatenated with the depth value, and then projected into 3D space via camera intrinsics and LiDAR-camera extrinsics to obtain the Gaussian's mean position $\boldsymbol{\mu}$ in the LiDAR frame.
Once the Gaussian embeddings are initialized, they are refined through a set of transformer~\cite{transformer} layers $\mathcal{F}$ that consists of a deformable attention module \cite{deformabledetr}, a self-attention module, and an MLP layer, as shown in Fig.~\ref{fig:method-arch}. The point $(u_i,v_i)$ serves as the reference point for deformable attention over the image features:
$$\mathbf{q_i^{\prime}} = \mathcal{F}(\mathbf{q_i}, (u_i, v_i)). \eqno{(2)}$$
Finally, the Gaussian parameters are predicted from embeddings $\mathbf{q_i}$ using an MLP head:
$$
g_i := \{\Delta\boldsymbol{\mu}_i,\;
        \mathbf{s}_i,\;
        \mathbf{r}_i,\;
        \alpha_i,\;
        \mathbf{f}_i\} = \operatorname{MLP_{GH}}(\mathbf{q_i^\prime}).
        \eqno{(3)}
$$
The means $\mathbf{\mu}$ are iteratively refined after each layer using the predicted offsets $\Delta\mathbf{\mu}$, while the scale $\mathbf{s}$ and opacity $\alpha$ are scaled to appropriate ranges. The rotations $\mathbf{r}$ are fixed to be axis-aligned in the camera's coordinate frame, and are not learned to simplify training.
The model is trained by differentiably rendering the DINOv2 features and depth from the predicted Gaussians using alpha-blending, implemented with optimized CUDA kernels \cite{ye2025gsplat}. The training objective penalizes the difference between the rendered outputs and corresponding predicted depth and features \cite{gausstr}.

\subsection{Joint Training}
\label{sec:joint-training}

GaussTR~\cite{gausstr} is trained \textit{per camera view}, with Gaussians predicted from each camera splatted back into that view for supervision. While this approach is effective for datasets with limited camera overlap, such as in the nuScenes dataset \cite{nuscenes}, it is suboptimal for multi-agent perception, where independent supervision fails to exploit overlapping observations across agents. 
Therefore, the model is trained \textit{per scene} as shown in Fig.~\ref{fig:method-arch}. All Gaussians predicted from cameras in a scene are aggregated and splatted into every camera in which they are visible, not only the cameras that produced them. Thus, predictions consistent with a single view but inconsistent with other views are penalized.
During inference, Gaussians are predicted from each camera, aggregated, and then used by agents for downstream tasks. CoGaussian can use weights from either per-image or per-scene training. In Sec.~\ref{sec:experiments}, we refer to the former as \textit{Zero-shot Concatenation}, which combines per-image training with our proposed inference method, and the later as \textit{Joint Training}, which uses per-scene training to enable collaboration.

\subsection{Radius-Gated Attention Fusion}
\label{sec:attention-fusion}

Joint Training improves performance as reported in Table~\ref{tab:main}. However, aggregating Gaussians from different cameras with overlapping fields of view causes the respective Gaussians to overlap. While this adds information for the ego, it also introduces noise. Additionally, the appearance of the same object often varies across different viewpoints. Thus, Gaussians for the same object may differ widely. Joint training helps mitigate such differences, but it does so \textit{implicitly} during Gaussian prediction. Therefore, we make this \textit{explicit} by introducing a \textit{fusion} module.
The fusion module aggregates useful information from Gaussians predicted by neighboring agents. We use single-head scaled dot product attention \cite{transformer} between ego Gaussian embeddings and neighboring Gaussian embeddings. Afterwards, a \textit{Fusion Head} with the same architecture as the Gaussian Head predicts Gaussians from the updated embeddings. The updated ego Gaussians and the neighboring Gaussians are aggregated to form the final set used for downstream tasks. The fusion pipeline is shown in Fig.~\ref{fig:method-arch}.

\textbf{Selecting neighbor Gaussians.} Let $\mathcal{G}_{ego}$ be the set of ego Gaussians and $\mathcal{G}_{ngh}$ the set of neighbor Gaussians, transformed into ego's coordinate frame. For each ego Gaussian $g_i \in \mathcal{G}_{ego}$, we take its $k$ nearest neighbors in $\mathcal{G}_{ngh}$ by Euclidean distance between Gaussian means, denoted by $\mathcal{N}(g_i)$. Then, we create an attention mask $\mathbf{M}$ based on distance between $g_i$ and $g_j \in \mathcal{N}(g_i)$:
Both $k$ and $\rho$ are hyperparameters. Taking the $k$ nearest neighbors imposes an intuitive locality bias: an ego gaussian can only be updated by nearby neighbor Gaussians. The bias is more strongly enforced by allowing Gaussians only within a specific radius $\rho$ to contribute to the update.

\textbf{Attention over Gaussian embeddings.} The fusion uses Gaussian embeddings obtained from the transformer layers on both ego and neighboring agents along with the optimized mean. The neighbor Gaussian embeddings and their means are communicated to the ego agent, where the attention matrix is computed:
$$
\Delta\boldsymbol{\mu}_{i,j} = \boldsymbol{\mu}_j - \boldsymbol{\mu}_i,
\quad
\mathbf{z}_{i,j} = \big[\, \mathbf{q}_j \;\|\; \Delta\boldsymbol{\mu}_{i,j} \;\|\;\operatorname{\eta}(\Delta\boldsymbol{\mu}_{i,j}) \,\big],
\eqno{(4)}
$$
$$
\mathbf{Q}_i = \phi_q(\mathbf{q}_i), \;\; \mathbf{K}_{i,j} = \phi_k(\mathbf{z}_{i,j}), \;\; \mathbf{V}_{i,j} = \phi_v(\mathbf{z}_{i,j}),
\eqno{(5)}
$$
$$
A_{i,j} = \frac{\langle \mathbf{Q}_i, \mathbf{K}_{i,j}\rangle}{\sqrt{d}}.
\eqno{(6)}
$$
Here, $\mathbf{q}_i$ is the embedding corresponding to Gaussian $g_i$. $\Delta\boldsymbol{\mu}_{i,j}$ is the relative position of $g_j$ with respect to $g_i$. $\phi_q$, $\phi_k$, and $\phi_v$ are two-layer MLPs and $\|$ is concatenation. The query vector is built from ego Gaussian embedding alone. The key and value vectors combine the neighbor Gaussian embeddings with their position relative to ego Gaussian along with its positional encoding $\operatorname{\eta}(\Delta\boldsymbol{\mu}_{i,j})$. $d$ is the dimension of $\mathbf{Q}_i$ and $\mathbf{K}_{i,j}$.

$$
w_{i,j} = \frac{\operatorname{softmax}_j(A_{i,j})\, M_{i,j}}{\sum_{j'} \operatorname{softmax}_{j'}(A_{i,j'})\, M_{i,j'} + \epsilon}, \eqno{(7)}
$$
$$
\mathbf{c}_i = \sum_{j \in \mathcal{N}(g_i)} w_{i,j}\, \mathbf{V}_{i,j}, \quad \hat{\mathbf{q}}_i = \mathbf{q}_i + \lambda\,\psi(\mathbf{c}_i). \eqno{(8)} 
$$
The attention mask is applied while calculating the softmax, which allows only radius-gated neighboring Gaussians to have non-zero weights $w_{i,j}$. An MLP $\psi$ is used to obtain an update which is scaled by $\lambda$ and added to the Gaussian embedding.

\textbf{Training.} We follow a two-stage training paradigm. Stage 1: We train the Transformer Layer and Gaussian Head as described in Sec. \ref{sec:joint-training}. Stage 2: The Transformer Layer and Gaussian Head are frozen, and the model learns the Radius-Gated Attention Layer and the Fusion Head.

\section{EXPERIMENTS}
\label{sec:experiments}
\definecolor{building}{HTML}{464646}
\definecolor{fence}{HTML}{8b5a3c}
\definecolor{terrain}{HTML}{8fb26a}
\definecolor{pole}{HTML}{f59e0b}
\definecolor{road}{HTML}{3b82f6}
\definecolor{sidewalk}{HTML}{ec4899}
\definecolor{vegetation}{HTML}{2f855a}
\definecolor{vehicles}{HTML}{dc2626}
\definecolor{wall}{HTML}{7c3aed}
\definecolor{guardrail}{HTML}{a16207}
\definecolor{trafficsigns}{HTML}{eab308}
\definecolor{bridge}{HTML}{14b8a6}

\begin{table*}[htbp]
    \centering
    \small
    \setlength{\tabcolsep}{4pt}

    \caption{3D Semantic Occupancy Prediction on Semantic-OPV2V Dataset} 
    \scalebox{0.94}{
    \makebox[\textwidth][c]{%
        \begin{tabular}{r|cc|c|ccc|c|ccc}
            \toprule
            
            \multirow{4}{*}{Method}
            & \multicolumn{6}{c|}{Fully-Supervised, Closed Vocabulary}
            & \multicolumn{4}{c}{Self-Supervised, Open Vocabulary} \\
            
            \cmidrule(lr){2-7}
            \cmidrule(lr){8-11}
            
            & \multicolumn{2}{c|}{Single-Agent}
            & \multicolumn{4}{c|}{Collaborative Perception}
            & \multicolumn{1}{c|}{Single-Agent}
            & \multicolumn{3}{c}{Collaborative Perception} \\
            
            \cmidrule(lr){2-3}
            \cmidrule(lr){4-7}
            \cmidrule(lr){8-8}
            \cmidrule(lr){9-11}
            
            & \multirow{2}{*}{CoHFF}
            & \multirow{2}{*}{GSFormer}
            & \multirow{2}{*}{CoHFF}
            & \multicolumn{3}{c|}{VOGS-CP}
            & \multirow{2}{*}{GaussTR}
            & \multicolumn{3}{c}{CoGaussian (Ours)} \\
            
            \cmidrule(lr){5-7}
            \cmidrule(lr){9-11}
            
            & & & &
            Zero Shot
            & Naive Fusion
            & Learned Fusion
            &
            & Concatenation
            & Trained
            & Attention \\

            \midrule
            IoU $(\uparrow)$        & 38.52  & \textbf{67.76} & 50.46 & 67.88 & \underline{70.10} & \textbf{72.87} & 20.58 & 16.13 & \underline{22.54} & \textbf{22.57}\\
            mIoU $(\uparrow)$        & 24.81  & \textbf{29.20} & 34.16 & 30.54 & \underline{36.02} & \textbf{37.44} & 6.26 & 4.81 & \underline{6.74} & \textbf{6.95}\\
            \midrule
            Building \textcolor{building}{$\blacksquare$}   & \textbf{21.04}  & 3.84  & \textbf{25.72}  & 3.91  & 7.18 & \underline{9.61} & \textbf{2.03} & 1.25 & 1.65 & \underline{1.93}\\
            Fence \textcolor{fence}{$\blacksquare$}      & \textbf{20.05}  & 14.10  & \underline{27.83} & 17.29 & 25.66 & \textbf{29.20} & 3.19 & 2.61 & \textbf{3.80} & \underline{3.39} \\
            Terrain \textcolor{terrain}{$\blacksquare$}    & 43.93  & \textbf{68.97} & 48.30 & 72.00 & \textbf{76.05} & \underline{74.51} & \underline{2.86} & 1.71 & 2.27 & \textbf{2.97} \\
            Pole \textcolor{pole}{$\blacksquare$}       & \textbf{31.66}  & 5.94 & \textbf{42.74}  & 8.44  & \underline{12.67} & 12.19 & 0.95 & 0.87 & \underline{0.96} & \textbf{1.04}\\
            Road \textcolor{road}{$\blacksquare$}       & 55.83 & \textbf{79.37}  & 61.77 & \underline{81.35} & 78.60 & \textbf{83.05} & 27.68 & 19.71 & \underline{32.41} & \textbf{32.70}\\
            Side walk \textcolor{sidewalk}{$\blacksquare$}  & 17.31 & \textbf{70.55}  & 39.62 & 68.91 & \underline{75.88} & \textbf{78.22} & \underline{14.90} & 12.57 & 13.62 & \textbf{17.29}\\
            Vegetation \textcolor{vegetation}{$\blacksquare$} & \textbf{14.49}  & 12.54  & \textbf{20.59} & 13.72 & 16.42 & \underline{20.43} & \underline{5.67} & 4.64 & \textbf{5.88} & 5.39\\
            Vehicles \textcolor{vehicles}{$\blacksquare$}   & \textbf{58.55}  & 49.25  & \textbf{63.28} & 49.56 & 57.14 & \underline{60.49} & 12.55 & 9.88 & \textbf{15.39} & \underline{14.24}\\
            Wall \textcolor{wall}{$\blacksquare$}       & \textbf{33.30}  & 30.79  & \textbf{58.27} & 30.49 & 35.25 & \underline{36.45} & \textbf{3.17} & \underline{2.83} & 2.49 & 2.53\\
            Guard rail \textcolor{guardrail}{$\blacksquare$} & 1.54  & \textbf{15.01}  & 1.94  & 20.78 & \textbf{41.00} & \underline{32.50} & \underline{0.31} & 0.16 & \textbf{0.85} & \underline{0.31}\\
            Traffic signs \textcolor{trafficsigns}{$\blacksquare$} & 0    & 0  & \textbf{16.33}     & 0     & 6.35 & \underline{8.26} & \underline{0.45} & 0.44 & \textbf{0.52} & \textbf{0.52}\\
            Bridge \textcolor{bridge}{$\blacksquare$}     & 0     & 0   & \underline{3.53}    & 0     & 0 & \textbf{4.35} & \textbf{1.33} & \underline{1.08} & 1.07 & 1.05 \\
            \bottomrule
        
        \end{tabular}
    }
    }
    \label{tab:main}
\end{table*}

\subsection{Dataset and Metrics}
\label{sec:dataset-n-metrics}

We evaluate on the Semantic-OPV2V benchmark~\cite{cohff}, an extension of OPV2V dataset~\cite{opv2v} with 3D SOP ground truth. OPV2V is a simulated dataset for vehicle-to-vehicle (V2V) collaboration with two to seven collaborating agents in a scene. Each vehicle is equipped with 4 cameras. The ground truth consists of 12 semantic labels over a $40 \times 40 \times 3.2m$ area with a grid size of $100 \times 100 \times 8$, giving a voxel size of $0.4m$.
Following previous work \cite{vogs-cp,cohff}, we report the per-class Intersection-over-Union (IoU), and its mean (mIoU), along with the overall occupancy IoU. Even though fully-supervised methods are included for reference, they are not comparable to our approach as they use dense 3D voxel supervision.

\subsection{Implementation Details}
\label{sec:implementation}

The model is trained per-view for 6 epochs for single-agent and zero-shot concatenation. Joint per-scene training is then run for 3 epochs. For Radius-Gated Attention Fusion, the transformer is frozen, and the attention module and fusion head are trained for 3 epochs. All training runs use a single NVIDIA A100 GPU and take 8-14 hours. The losses converge within 2 epochs for all runs. Metric3Dv2~\cite{metric3dv2} provides depth supervision, and DINOv2~\cite{dinov2} provides open-vocabulary feature supervision. During inference, Gaussian features are compared with text embeddings aligned to DINOv2 using Talk2DINO~\cite{barsellotti2025talkingdinobridgingselfsupervised}, prompted with the 12 classes plus an extra class representing free space. We use AdamW with a learning rate of $1 \times 10^{-5}$, weight decay of $5 \times 10^{-3}$, and gradient clipping of 20. We evaluate models using 1200 Gaussian queries per camera in four configurations:
\begin{itemize}
\item \textbf{Single Agent Perception:} Each agent independently predicts occupancy.

\item \textbf{Zero-shot Concatenation:} Gaussians predicted by all agents are aggregated in a common coordinate frame.

\item \textbf{Joint Training:} Aggregated Gaussians from all cameras are used for fine-tuning.

\item \textbf{Radius-Gated Attention:} The fusion pipeline runs for each agent, updating ego Gaussians using neighboring Gaussians.

\end{itemize}

\section{RESULTS AND DISCUSSION}
\label{sec:results}

\subsection{Quantitative Results}
Table \uppercase\expandafter{\romannumeral 1} highlights quantitative results from previous works, as well as our experiments. Relative to the single-agent GaussTR baseline, zero-shot concatenation reduces performance by $\mathbf{-4.45}$/$\mathbf{-1.45}$ (IoU/mIoU), showing that naive Gaussian aggregation introduces spatially inconsistent structure. Joint Training improves over the baseline by $\mathbf{+1.96}$/$\mathbf{+0.48}$ (IoU/mIoU), indicating that cross-view rendering supervision reconciles these inconsistencies. Using Radius-Gated Attention gives a modest gain of $\mathbf{+1.99}$/$\mathbf{+0.69}$ (IoU/mIoU) over the baseline.

Class-wise behavior is mixed, with frequently observed classes like ``road", ``sidewalk", and ``vehicles" benefiting the most from collaboration, as they have major overlap among agents and thus a stronger supervision signal. Prominent but less visible classes like "vegetation" and "terrain" struggle with collaboration and other small and uncommon classes have mixed results.

\subsection{Qualitative Results}

Fig. \ref{fig:visualisation} shows that the collaborative methods recover structure outside the ego view. Our method tends to produce volumetric predictions rather than one voxel thick surfaces in the ground truth. This reflects the nature of sparse 3D Gaussian representations, unlike methods like \cite{gaussianformer, gaussianformer2, zhao2025gaussianformer3d} that rely on 3D semantic voxel supervision and many surface-fitting small Gaussians. 2D image-based supervision also makes objects occupying large image regions more prominent in 3D space. The elongated prediction of Vehicle~1 and volumetric prediction of Vehicle~2 demonstrate this behavior.

Joint Training reduces these inconsistencies, producing smaller and more conservative Gaussians, evident in Vehicle~2 and smaller elliptical predictions throughout the scene. The fusion pipeline preserves this coherence while recovering more dominant regions of the scene, including the Road and some Vehicles. However, thin and infrequent structures remain difficult to reconstruct reliably under purely 2D self-supervision.

\section{CONCLUSION AND FUTURE WORK}

We introduce CoGaussian which is, to the best of our knowledge, the first framework for open-vocabulary collaborative perception that communicates foundation model-aligned 3D Gaussians among agents to represent a scene. Evaluation shows that collaboration is effective when interaction is learned. Zero-shot collaboration hurts performance, whereas joint training and radius-gated attention improve over the single-agent baseline. CoGaussian also brings the flexibility of open-vocabulary to collaborative perception.

We plan to explore different fusion architectures and evaluate additional downstream tasks such as BEV Semantic Segmentation and 3D Object Detection. Efficient communication and inference optimization will also be crucial for enabling real-time implementation.

\addtolength{\textheight}{-12cm}   %

{\scriptsize
\bibliographystyle{IEEEtran}
\bibliography{references}
}

\end{document}